\documentclass[sigconf]{acmart}

\AtBeginDocument{%
  \providecommand\BibTeX{{%
    \normalfont B\kern-0.5em{\scshape i\kern-0.25em b}\kern-0.8em\TeX}}}

\setcopyright{rightsretained}
\copyrightyear{2022}
\acmYear{2022}
\acmConference{SIGGRAPH '22 Immersive Pavilion }{August 07-11, 2022}{Vancouver, BC, Canada}
\acmBooktitle{Special Interest Group on Computer Graphics and Interactive Techniques Conference Immersive Pavilion (SIGGRAPH '22 Immersive Pavilion ), August 07-11, 2022}
\acmDOI{10.1145/3532834.3536210}
\acmISBN{978-1-4503-9369-0/22/08}

\usepackage[subrefformat=parens]{subcaption}
\usepackage{algorithm}
\usepackage{algorithmic}

\begin{document}

\title{Immersive Real World through Deep Billboards}

\author{Naruya Kondo}

\orcid{0000-0002-9694-4676}
\affiliation{%
  \institution{\mbox{n-kondo@digitalnature.slis.tsukuba.ac.jp}}
  \country{University of Tsukuba, Japan}
}

\author{So Kuroki}
\email{kuroki-so273@g.ecc.u-tokyo.ac.jp}
\affiliation{%
  \country{The University of Tokyo, Japan}
}

\author{Ryosuke Hyakuta}
\email{momosuke@digitalnature.slis.tsukuba.ac.jp}
\affiliation{%
  \country{University of Tsukuba, Japan}
}

\author{Yutaka Matsuo}
\email{matsuo@weblab.t.u-tokyo.ac.jp}
\affiliation{%
  \country{The University of Tokyo, Japan}
}

\author{Shixiang Shane Gu}
\email{shanegu@google.com}
\affiliation{%
  \country{Google Brain, USA}
}

\author{Yoichi Ochiai}
\email{wizard@slis.tsukuba.ac.jp}
\affiliation{%
  \country{University of Tsukuba, Japan}
}

\renewcommand{\shortauthors}{Naruya Kondo, et al.}

\begin{abstract}
An aspirational goal for virtual reality (VR) is to bring in a rich diversity of real world objects \textit{losslessly}. Existing VR applications often convert objects into explicit 3D models with meshes or point clouds, which allow fast interactive rendering but also severely limit its quality and the types of supported objects, fundamentally upper-bounding the “realism” of VR. Inspired by the classic “billboards” technique in gaming, we develop Deep Billboards that model 3D objects \textit{implicitly} using neural networks, where only 2D image is rendered at a time based on the user’s viewing direction. Our system, connecting a commercial VR headset with a server running neural rendering, allows real-time high-resolution simulation of detailed rigid objects, hairy objects, actuated dynamic objects and more in an interactive VR world, drastically narrowing the existing real-to-simulation (real2sim) gap. Additionally, we augment Deep Billboards with physical interaction capability, adapting classic billboards from screen-based games to immersive VR.  At our pavilion, the visitors can use our off-the-shelf setup for quickly capturing their favorite objects, and within minutes, experience them in an immersive and interactive VR world – with minimal loss of reality. Our project page: \url{https://sites.google.com/view/deepbillboards/}
\end{abstract}



\keywords{image based rendering, neural networks, billboard}

\begin{teaserfigure}
  \centering
  \includegraphics[width=0.85\textwidth]{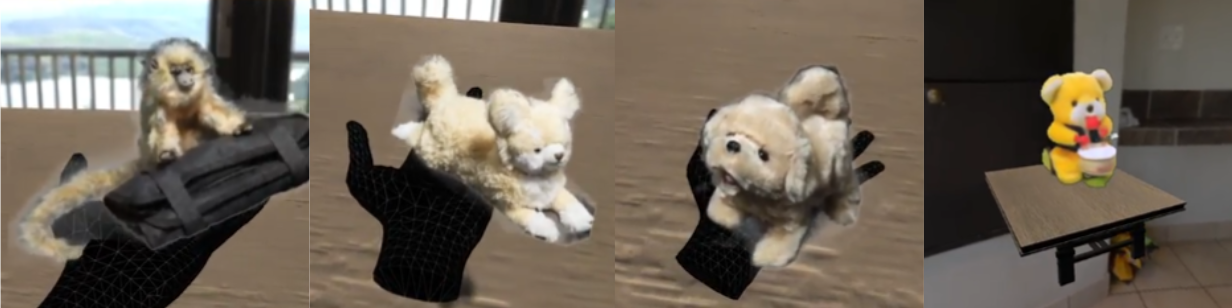}
  \caption{Interactive Deep Billboards with a NeRF model (left three) and a World model (right most) in VR.}
  \label{fig:teaser}
\end{teaserfigure}

\maketitle

\section{Introduction}

Data-driven 3D modeling is increasingly in demand from the VR industry, and is essential for reality-grounded VR applications such as shopping, showrooms, tourism, and teleconferences that require transporting real objects into the interactive virtual world. Traditionally, VR uses explicit 3D models such as meshes and point clouds for fast simulation, but they have limited rendering quality and thus fundamentally upper-bound the “realism” of current VR worlds. Inspired from classic ``2D'' billboards in gaming, we propose Deep Billboards and develop a novel system that drastically enhances real-to-simulation (real2sim) quality of complex objects in interactive VR.

\section{Deep Billboards}

    

\begin{figure}[htbp]
  \def \factor {0.31}  
  \def \vertspace {0cm}
  \def \horizontalspace {2pt}
  \centering
  \begin{tabular}{c@{\hspace{\horizontalspace}}c@{\hspace{\horizontalspace}}c@{\hspace{\horizontalspace}}c@{\hspace{\horizontalspace}}c}
    \includegraphics[width=0.35\linewidth]{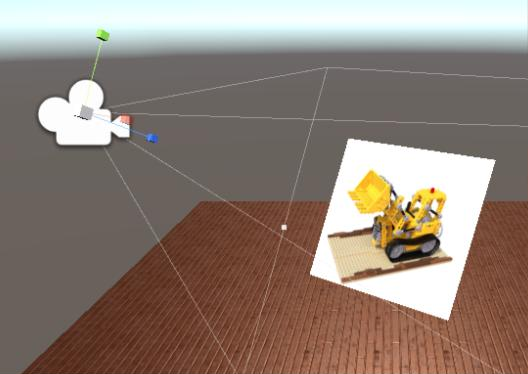} &
    \includegraphics[trim=77px 0px 77px 0px, clip, width=0.245\linewidth]{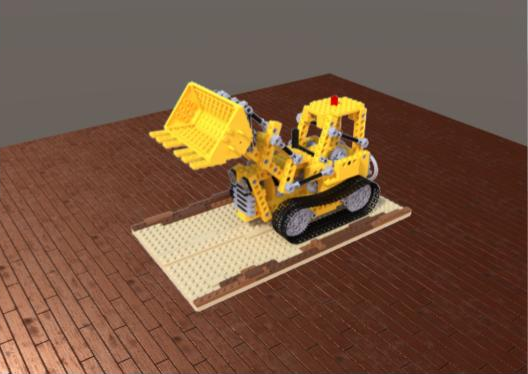} &
    \includegraphics[width=0.35\linewidth]{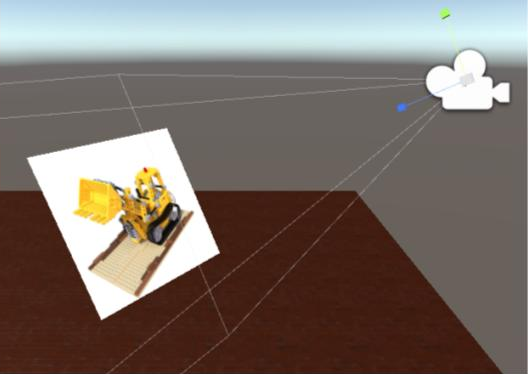}
    \vspace{\vertspace}
  \end{tabular}
  \caption{Deep Billboard object on a single canvas.}
  \label{fig:camera}

\end{figure}

\begin{figure}[htbp]
  \def \factor {0.44} 
  \def \factorb {0.29} 
  \def \vertspace {0cm}
  \def \horizontalspace {2pt}
  \centering
  \begin{minipage}[t]{0.98\hsize}
  \begin{tabular}{c@{\hspace{\horizontalspace}}c@{\hspace{\horizontalspace}}c}
    \includegraphics[width=\factor\linewidth]{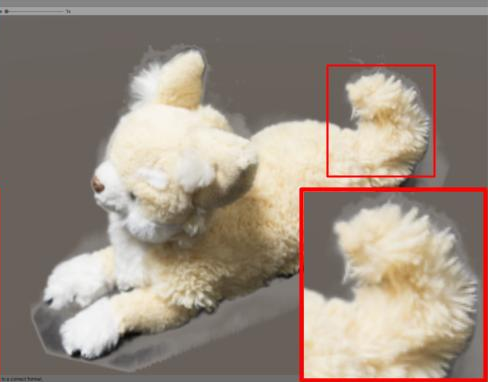} &
    \includegraphics[width=\factor\linewidth]{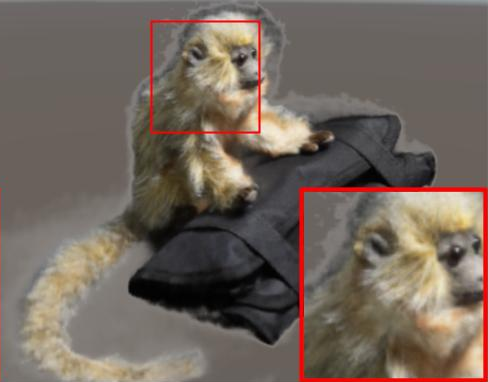}
  \end{tabular}

  \end{minipage}
  \begin{minipage}[t]{0.98\hsize}
  \begin{tabular}{c@{\hspace{\horizontalspace}}c@{\hspace{\horizontalspace}}c@{\hspace{\horizontalspace}}c}
    \includegraphics[width=\factorb\linewidth]{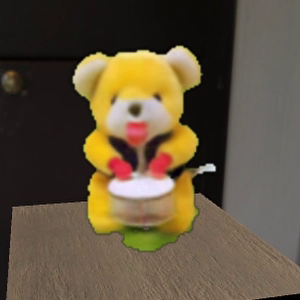} &
    \includegraphics[width=\factorb\linewidth]{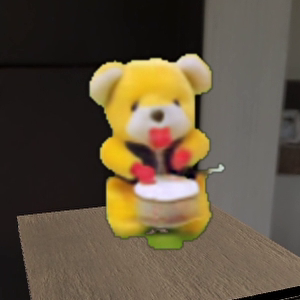} &
    \includegraphics[width=\factorb\linewidth]{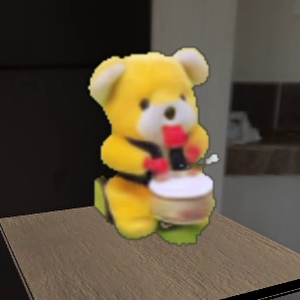}
  \end{tabular}
  \end{minipage}


  \caption{Quality of NeRF Billboard (top) and dynamic rendering of World Billboard (bottom).}
  \label{fig:db-nerf}

\end{figure}

A ``billboard'' in computer graphics and 3D video games corresponds to a 2D image that is tilted to always face the observer, creating an illusion of a 3D object. 
 We propose Deep Billboard, a generalization of the classic billboard, where the 2D texture is re-generated at each frame from an implicit neural rendering model outputting the view-conditioned images (Figure. \ref{fig:camera}). Compared to standard explicit 3D models such as meshes or point clouds used in VR, Deep Billboards enable object rendering of much higher resolutions, drastically improving the realism of virtual experiences while preserving real-time interactivity. Any implicit neural rendering model can be used for our Deep Billboard (Figure. \ref{fig:db-nerf}).

\paragraph{NeRF Billboard}
Neural randiance field (NeRF) has revolutionized the field of data-driven 3D reconstruction ~\citep{mildenhall2020nerf}.
While NeRF allows a much higher-resolution 3D construction than classic approaches, and requiring only tens of 2D images of the object from multiple angles. Prior work importing NeRF into VR converting the models back to coarse meshes, due to software incompatibility, high computational demand, and slow rendering of original NeRF models. 
Our system instead used PlenOctrees \cite{yu2021plenoctrees}, an extension of NeRF to allow real-time rendering, running on a remote GPU server, to directly use NeRF to update the billboards. To the best of our knowledge, our system is the first to import NeRF models directly into an interactive VR application without loss of accuracy.

\paragraph{World Billboard}
While NeRF allows rich rendering of static objects, extension to dynamic scenes is still an active area of research. To show the versatility of our system, we also replace our billboard updater using PlaNet~\cite{hafner2019planet}, a neural state-space model (SSM) or a \textit{world} model for action-conditioned video prediction in deep reinforcement learning (RL), to achieve data-driven reproduction of time-varying objects from a single video only. From a single 10-minute video labeled with the camera position and orientation, we learned video prediction conditioned on the initial viewpoint image and viewpoint series, and reproduced the time-varying 3D object through the billboard in our interactive VR system. 

\paragraph{Physics Interaction}
\begin{figure}[htbp]
    \def \factor {0.3} 
  \def \horizontalspace {2pt}
    \begin{tabular}{c@{\hspace{\horizontalspace}}c@{\hspace{\horizontalspace}}c@{\hspace{\horizontalspace}}}
        \begin{minipage}[t]{\factor\hsize}
            \centering
            \includegraphics[width=\linewidth]{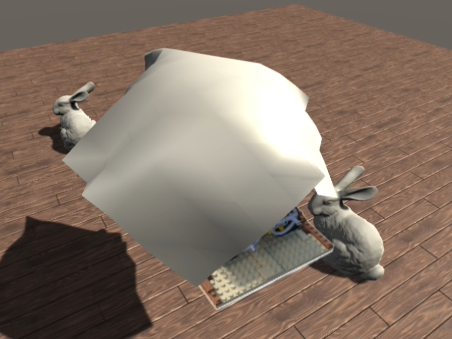}
            \subcaption{Cloths}
            \label{fig:clothes}
        \end{minipage} &
        \begin{minipage}[t]{\factor\hsize}
            \centering
            \includegraphics[width=\linewidth]{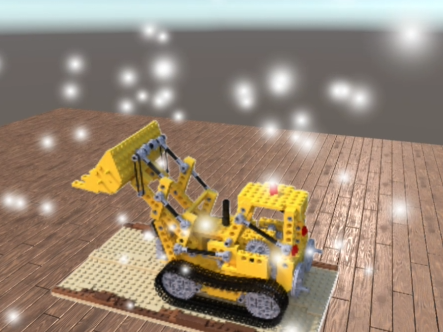}
            \subcaption{Snow}
            \label{fig:snow}
        \end{minipage} &
        \begin{minipage}[t]{\factor\hsize}
            \centering
            \includegraphics[width=\linewidth]{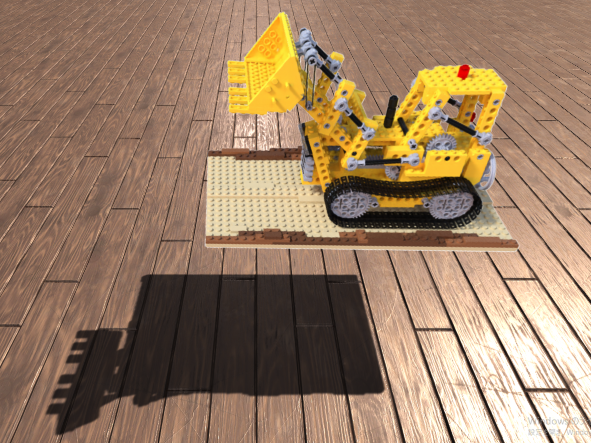}
            \subcaption{Shadow}
            \label{fig:shadow}

        \end{minipage}

    \end{tabular}
    \caption{Deep Billboard provides basic physical interaction.}
    \label{fig:phys}
\end{figure}


Another disadvantage of a classic billboard is that it does not allow physical interaction. To solve this problem, we propose a system where a billboard is used for visual interaction while an invisible rough mesh for physics interaction. In Figure ~\ref{fig:phys}, we used a mesh extracted from a NeRF model and realized three basic physical interactions. This makes our Deep Billboard to achieve both good visuals and quality physics. 

\section{System}
\begin{figure}[htbp]
  \centering
  \includegraphics[width=0.9\linewidth]{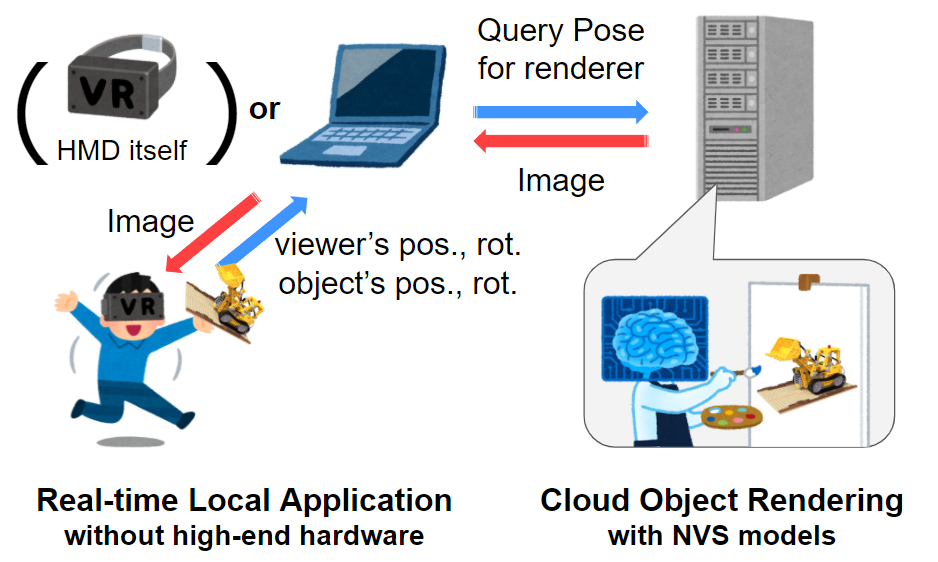}
  \caption{Deep Billboard VR system overview.}
  \label{fig:system}
\end{figure}
Deep generative models require large computational power, so we have built a system that uses cloud rendering to enable even inexpensive VR headsets to handle highly accurate pseudo-3D models as long as they have an internet connection (Figure. \ref{fig:system}). Each Deep Billboard object is rendered on a cloud server, and rendered 2D frame is sent to the VR system per frame to achieve real-time interaction with minimal on-board processing. Our system transported a wide range of objects, including actuated toys, hairy plushy to the interactive virtual world with minimal loss of reality.

\begin{acks}
This work was supported by the Strategic Information and Communications R\&D Promotion Programme (SCOPE) of the Ministry of Internal Affairs and Communications of Japan and New Energy and Industrial Technology Development Organization (NEDO) of Ministry of Economy, Trade and Industry of Japan.
\end{acks}

\bibliographystyle{format/ACM-Reference-Format}
\bibliography{base}



\end{document}